\documentclass{article}

\usepackage[a4paper, total={5.8in, 8in}]{geometry}
\usepackage{url}
\usepackage[numbers]{natbib}
\usepackage{amsmath}
\usepackage{titlesec}
\usepackage{authblk}

\titleformat*{\section}{\large\bfseries}
\titleformat*{\subsection}{\normalsize\bfseries}
\titleformat*{\subsubsection}{\normalsize\bfseries}
\titleformat*{\paragraph}{\normalsize\bfseries}
\titleformat*{\subparagraph}{\normalsize}

\begin{document}
    \title{\Large{Bounds of \textbf{MIN\_NCC} and \textbf{MAX\_NCC} and filtering scheme for graph domain variables}}
    
    \author[,1,2,3]{Dimitri Justeau-Allaire\thanks{Corresponding author: \texttt{dimitri.justeau@gmail.com}}}
    \author[1,2,3]{Philippe Birnbaum}
    \author[4]{Xavier Lorca}
    
	\affil[1]{CIRAD, UMR AMAP, Montpellier, France}
	\affil[2]{Institut Agronomique néo-Calédonien (IAC), Noum\'ea, New Caledonia}
	\affil[3]{AMAP, Univ Montpellier, CIRAD, CNRS, INRAE, IRD, Montpellier, France}
	\affil[4]{Centre de G\'enie Industriel, IMT Mines Albi, Albi, France}

    \date{May 2018}
    \maketitle{}

    \begin{abstract}
        Graph domain variables and constraints are an extension of constraint programming introduced by \citeauthor{dooms_2005} \cite{dooms_2005}. This approach had been further investigated by \citeauthor{fages_2015} in its PhD thesis \cite{fages_2015}. On the other hand, \citeauthor{beldiceanu_2006} presented a generic filtering scheme for global constraints based on graph properties \cite{beldiceanu_2006}. This scheme strongly relies on the computation of graph properties' bounds and can be used in the context of graph domain variables and constraints with a few adjustments \cite{beldiceanu_2006a}. Bounds of \textbf{MIN\_NCC} and \textbf{MAX\_NCC} had been defined for the graph-based representation of global constraint for the \texttt{path\_with\_loops} graph class \cite{beldiceanu_2006}. In this note we generalize those bounds for graph domain variables and for any graph class. We also provide a filtering scheme for any graph class and arbitrary bounds.
    \end{abstract}

    \section{Introduction and Notations}

    We use the notations introduced in \cite{beldiceanu_2006}.

    \begin{itemize}
        \item $G$: The graph domain variable.
        \item $V_T$: The set of mandatory vertices.
        \item $E_T$: The set of mandatory arcs.
        \item $V_U$: The set of potential vertices.
        \item $E_U$: The set of potential arcs.
        \item $V_F$: The set of removed vertices.
        \item $E_F$: The set of removed arcs.
        \item $G(V_T, E_T)$: The graph kernel.
        \item $G(V_{TU}, E_{TU})$: The graph envelope.
        \item $cc(G)$: The set of connected components of $G$, $cc_{[cond]}(G)$ denotes the set of connected components of $G$ verifying $cond$.
        \item If $i$ is a connected component, $l_i$ is the length of any elementary sequence made from all the vertices of $i$, or more simple, $l_i$ is the size of $i$.
    \end{itemize}

    \section{The MIN\_NCC Constraint}

    Given a graph domain variable $G$ and an integer domain variable $P$, the constraint $\text{MIN\_NCC}(G, P)$ holds if $P$ is the size of the smallest connected component of $G$. We denote by $\underline{\text{MIN\_NCC}}(G)$ and $\overline{\text{MIN\_NCC}}(G)$ the lower and upper bounds of the $\text{MIN\_NCC}$ graph property, computed from $G$. Sharp bounds of $\text{MIN\_NCC}$ had been defined in \cite{beldiceanu_2006} for graph-based representation of global constraints, in the particular case of the \texttt{path\_with\_loops} graph class. We adjusted them for the context of graph domain variables, and for any graph class.

    \begin{align}
        &\underline{\text{MIN\_NCC}}(G) = \left\{
        \begin{array}{ll}
            0 & \text{if } |V_T| = 0; \\
            1 & \text{if } |V_T| \geq 1 \land |V_U| \geq 1; \\
            \min_{i \in \textit{cc}(G(V_T, E_T))} l_i & \text{if } |V_T| \geq 1 \land |V_U| = 0.
        \end{array}
        \right. \\[6pt]
        &\overline{\text{MIN\_NCC}}(G) = \left\{
        \begin{array}{ll}
            \min_{i \in \textit{cc}_{[|V_T| \geq 1]}(G(V_{TU}, E_{TU}))} l_i & \text{if } |V_T| \geq 1; \\
            \max_{i \in \textit{cc}(G(V_{TU}, E_{TU}))} l_i & \text{if } |V_T| = 0.
        \end{array}
        \right.
    \end{align}

    We now suggest a filtering scheme for the $\text{MIN\_NCC}(G, P)$ constraint that is adapted from the filtering scheme introduced in \cite{beldiceanu_2006} for the \texttt{path\_with\_loops} graph class in the context of graph-based representation of global constraints.

    \begin{enumerate}
        \item \textit{Trivial case}: if $\underline{P} > |V_{TU}|$ then \textbf{\textit{fail}}.
        \item \textit{Trivial case}: if $\underline{P} > 0$ and $|E_{TU}| < \max(1, \underline{P} - 1)$ then \textbf{\textit{fail}}.
        \item If $\overline{\text{MIN\_NCC}}(G) < \underline{P}$ then \textbf{\textit{fail}}. Note that a sufficient condition is $|\textit{cc}_{[|V_{TU}| < \underline{P} \land |V_T| \geq 1]}(G(V_{TU}, E_{TU}))| \geq 1$, that is the existence of a mandatory connected component in the envelope strictly smaller than $\underline{P}$.
        \item\label{step:fail_inf_ge_sup} If $\underline{\text{MIN\_NCC}}(G) > \overline{P}$ then \textbf{\textit{fail}}.
        \item\label{step:set_p_inf} If $\underline{P} < \underline{\text{MIN\_NCC}}(G)$ then set $\underline{P}$ to $\underline{\text{MIN\_NCC}}(G)$.
        \item If $\overline{P} > \overline{\text{MIN\_NCC}}(G)$ then set $\overline{P}$ to $\overline{\text{MIN\_NCC}}(G)$.
        \item Every $U$-vertex of any optional connected component smaller than $\underline{P}$ (that is, in $\textit{cc}_{[|V_{TU}| < \underline{P} \land |V_T| = 0]}\allowbreak(G(V_{TU}, E_{TU}))$) is turned into a $F$-vertex.
        \item If $\underline{\text{MIN\_NCC}}(G) < \underline{P}$ then:
        \begin{enumerate}
            \item If $|V_T| \geq 1$, every $U$-vertex of any mandatory connected component of size $\underline{P}$ (that is, in $\textit{cc}_{[|V_{TU}| = \underline{P} \land |V_T| \geq 1]}(G(V_{TU}, E_{TU}))$), is turned into a $T$-vertex. $P$ can then be instantiated to $\underline{P}$.
            \item\label{step:inf_min_ncc_le_inf_p} Consider each connected component $i \in \textit{cc}_{[|V_{T}| < \underline{P}]}(G(V_{T}, E_{T}))$, if there is exactly one $U$-arc ($x$, $y$) outgoing from $i$ ($x \in i \land y \notin i$), it is turned into a $T$-arc. If $y$ is a $U$-vertex, it is turned into a $T$-vertex. Similarly, if there are several $U$-arcs outgoing from $i$ but all of them are connected to a single $U$-vertex, then it is turned into a $T$-vertex.
        \end{enumerate}
        \item If $G$ had been modified in step \ref{step:inf_min_ncc_le_inf_p}, recompute $\underline{\text{MIN\_NCC}}(G)$ and repeat steps \ref{step:fail_inf_ge_sup} and \ref{step:set_p_inf}.
        \item If $\overline{\text{MIN\_NCC}}(G) > \overline{P}$ then:
        \begin{enumerate}
            \item If $\overline{P} = 0$ every $U$-vertex is turned into a $F$-vertex.
            \item If $\overline{P} = 1 \land |V_U| = 1 \land \min_{i \in \textit{cc}(G(V_T, E_T))} l_i > 1$ then the only $U$-vertex is turned into a $T$-vertex and every $U$-arc outgoing from it and which is not a loop is turned into a $F$-arc.
        \end{enumerate}
    \end{enumerate}

    \section{The MAX\_NCC Constraint}

    Given a graph domain variable $G$ and an integer domain variable $P$, the constraint $\text{MAX\_NCC}(G, P)$ holds if $P$ is the size of the largest connected component of $G$. We denote by $\underline{\text{MAX\_NCC}}(G)$ and $\overline{\text{MAX\_NCC}}(G)$ the lower and upper bounds of the $\text{MAX\_NCC}$ graph property, computed from $G$. Sharp bounds of $\text{MAX\_NCC}$ had been defined in \cite{beldiceanu_2006} for graph-based representation of global constraints, in the particular case of the \texttt{path\_with\_loops} graph class. Those bounds remain unchanged in the context of graph domain variables, for any graph class.

    \begin{align}
        &\underline{\textbf{MAX\_NCC}}(G) = \max_{i \in \textit{cc}(G(X_T, E_T))} l_i; \\
        &\overline{\textbf{MAX\_NCC}}(G) = \max_{i \in \textit{cc}(G(X_{TU}, E_{TU}))} l_i.
    \end{align}

    We now suggest a filtering scheme for the $\text{MAX\_NCC}(G, P)$ constraint that is adapted from the filtering scheme introduced in \cite{beldiceanu_2006} for the \texttt{path\_with\_loops} graph class in the context of graph-based representation of global constraints.

    \begin{enumerate}
        \item \textit{Trivial case}: if $\underline{P} > |X_{TU}|$ then \textbf{\textit{fail}}.
        \item \textit{Trivial case}: if $\underline{P} > 0$ and $|E_{TU}| < \max(1, \underline{P} - 1)$ then \textbf{\textit{fail}}.
        \item\label{step:sup} If $\overline{\textbf{MAX\_NCC}}(G) < \underline{P}$ then \textbf{\textit{fail}}.
        \item If $\underline{\textbf{MAX\_NCC}}(G) > \overline{P}$ then \textbf{\textit{fail}}. Note that a sufficient condition is $|\textit{cc}_{[|X_{T}| > \overline{P}]}(G(X_{T}, E_{T}))| \geq 1$, that is the existence of connected component in the kernel strictly greater than $\overline{P}$.
        \item If $\underline{P} < \underline{\text{MAX\_NCC}}(G)$ then set $\underline{P}$ to $\underline{\text{MAX\_NCC}}(G)$.
        \item\label{step:set_p_sup} If $\overline{P} > \overline{\text{MAX\_NCC}}(G)$ then set $\overline{P}$ to $\overline{\text{MAX\_NCC}}(G)$.

        \item If $\overline{\text{MAX\_NCC}}(G) > \overline{P}$ then:
        \begin{enumerate}
            \item If $\overline{P} = 1$ then every $U$-arc $(x, y)$ such that $x \neq y$ is turned into a $F$-arc.
            \item If $\overline{P} = 0$ then every $U$-vertex is turned into a $F$-vertex.
            \item Every $U$-arc linked to a maximal size connected component in the kernel, that is in $\textit{cc}_{[|X_{T}| = \overline{P}]}(G(X_{T}, E_{T}))$ is turned into a $F$-arc.
            \item\label{step:cut_cc} Any $U$-arc linking two connected components of $G(X_T, E_T)$ of size $s_1$ and $s_2$ such that $s_1 + s_2 > \overline{P}$ is turned into a $F$-arc.
            \item If $G$ had been modified in step \ref{step:cut_cc}, recompute $\overline{\text{MAX\_NCC}}(G)$ and repeat steps \ref{step:sup} and \ref{step:set_p_sup}.
        \end{enumerate}
        \item  If $|\textit{cc}_{[|V_{TU}| \geq \underline{P}]}(G(V_{TU}, E_{TU}))| = 1$ and the size of this single candidate component is exactly $\underline{P}$ then any $U$-vertex of it is turned into a $T$-vertex. $P$ can then be instantiated to $\underline{P}$.
    \end{enumerate}

    \bibliographystyle{plainnat}
    \bibliography{bibliography}
\end{document}